\documentclass{article}
\usepackage{spconf,amsmath,graphicx,hyperref}
\usepackage{array}
\usepackage{amsthm}
\usepackage{amssymb}
\usepackage{comment}
\usepackage{multirow}
\usepackage{xcolor}
\usepackage{algorithm}
\usepackage{algorithmic}
\usepackage{booktabs}
\usepackage{microtype}
\usepackage{subfigure}
\usepackage{adjustbox}
\usepackage{hyperref}
\usepackage{array}
\usepackage{multirow} 

\title{Ister: Linear Transformer for Efficient Multivariate \\ Time Series Forecasting}
\name{Fanpu Cao$^{1,4*}$, Shu Yang$^2$, Zhengjian Chen$^3$, Ye Liu$^4$, Laizhong Cui$^2$
\thanks{*Corresponding author: fanpucao@gmail.com}
\thanks{Thanks to the Red Bird MPhil Program at the Hong Kong University of Science and Technology (Guangzhou) for providing resources and funding.}
}
\address{$^1$Hong Kong University of Science and Technology (Guangzhou), Guangzhou, China \\ $^2$ShenZhen University, $^3$Shenzhen Energy Group Co., Ltd., Shenzhen, China \\ $^4$South China University of Technology, Guangzhou, China}
\begin{document}
\ninept
\maketitle
\begin{abstract}
Transformer-based models have achieved remarkable success in multivariate time series forecasting (MTSF) by capturing long-range dependencies. However, their widespread adoption is hindered by the quadratic computational complexity of self-attention, which limits scalability on high-dimensional sequences. To address this challenge, we propose the \textit{Inverted Seasonal-Trend Decomposition Transformer (Ister)}, a novel architecture that enhances both predictive accuracy and computational efficiency. Central to Ister is \textit{Dot-attention}, a linear-complexity attention mechanism that replaces conventional multi-head self-attention with element-wise dot-product operations to model inter-series dependencies. Furthermore, we introduce an inverted seasonal-trend decomposition strategy that isolates periodic components, enabling the model to focus learning on periodic patterns, thereby improving the performance of channel alignment. Extensive experiments across several real-world benchmarks demonstrate that Ister consistently achieves state-of-the-art performance. Code is available at https://github.com/macovaseas/Ister.
\end{abstract}
\section{Introduction}
\label{sec:intro}

Multivariate time series forecasting (MTSF) is widely applied in energy~\cite{energy}, transportation~\cite{trafficflow}, weather prediction~\cite{tcn} and economic planning~\cite{economic}, and accurate prediction is crucial for these domains~\cite{survey}. Recent work~\cite{iTransformer} emphasizes the critical role of modeling inter-series dependencies for MTSF performance. Accurately capturing cross-variable dependencies is crucial for forecasting performance, as the predictive accuracy of multivariate time series fundamentally hinges on modeling these intricate inter-series dependencies.

\begin{figure}[t]
  \centering
  \includegraphics[width=0.45\textwidth]{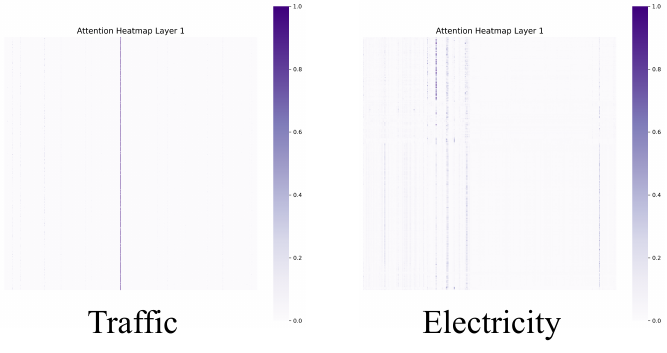}
  \caption{Attention heatmap learned by iTransformer~\cite{iTransformer} after 10 epochs of training on ECL and traffic dataset.}
  \label{fig:heatmap}
\end{figure}

Recent Transformer-based models have significantly advanced MTSF through self-attention mechanisms that capture long-range dependencies~\cite{informer}. State-of-the-art approaches like iTransformer~\cite{iTransformer} and TimeXer~\cite{TimeXer} leverage self-attention to model multivariate correlations, underscoring the necessity of cross-channel information transfer. However, they face significant scalability challenges when processing high-dimensional sequences due to the quadratic computational complexity of self-attention. At the same time, accurately identifying the correlation patterns among multivariate input sequences is inherently challenging. As illustrated in Fig. \ref{fig:heatmap}, the attention heat map learned by iTransformer on two well-known datasets~\cite{autoformer} displays a \textit{sparse stripe}, suggesting that only a few channels are strongly correlated with the target. In real-world scenarios with numerous channels, many contain substantial noise or lack meaningful semantic content. Processing all channels via full self-attention is computationally prohibitive and introduces irrelevant information that degrades forecasting performance. Thus, efficiently capturing channel-wise dependencies is paramount for building effective MTSF models.

To tackle the above mentioned problem, we propose a new Transformer-based model with linear complexity named \textbf{I}nverted \textbf{s}easonal-\textbf{t}rend decomposition Transform\textbf{er} (\textit{Ister}). Particularly, \textit{Ister} firstly decomposes the time series into seasonal and trend components, apply inverted-embedding on time points and output the variable tokens for each dimension. Secondly, \textit{Ister} replaces the multi-head self attention module with \textit{Dot-attention} to reduce computation complexity and uses Transformer encoder to forecast the seasonal component. Above all, \textit{Ister} effectively models multivariate correlations while maintaining variable independence. To evaluate the performance of \textit{Ister}, we conduct comprehensive simulations on different datasets. The results show that \textit{Ister} achieves state-of-the-art performance on real-world benchmarks and outperforms existing Transformer-based forecasters on long-term forecasting.

To sum up, the main contributions of this work include:
\begin{itemize}
    \item We propose \textit{Ister}, a novel linear Transformer model that efficiently models multivariate correlations for MSTF.
    \item We propose \textit{Dot-attention}, which replaces matrix multiplication with element-wise multiplication in self-attention, reducing computational complexity from quadratic to linear.
    \item We conduct extensive experiments on predicting long multivariate sequences on several real-world benchmarks, and prove that \textit{Ister} outperforms previous methods and achieves state-of-the-art performance.
\end{itemize}

\section{Related Work}
\label{sec:related work}

Transformer~\cite{transformer} has achieved remarkable success in NLP and gained significant attention for time series prediction, due to its ability to capture long-range dependencies. Previous works~\cite{reformer, informer, autoformer, fedformer} are devoted to reducing quadratic computational complexity in both computation and memory of the vanilla Transformer, while improving its prediction accuracy. For instance, Reformer~\cite{reformer} uses locality-sensitive hashing to improve the efficiency of Transformers. Informer~\cite{informer} introduces \textit{ProbSparse} self-attention to reduce the quadratic complexity in both time and memory domains. However, the point-wise representation performs poorly in capturing local semantics in temporal variations, ultimately leading to a decline in forecasting performance. At the same time, as the sequence length increases, the inference time of Transformer models grows significantly, limiting their practicality. 

To improve it, PatchTST~\cite{PatchTST} divides time series data into subseries-level patches and use self-attention to capture dependencies between these patches. Crossformer~\cite{Crossformer} segments each dimension of the time series to capture temporal and cross-dimensional dependencies. iTransformer~\cite{iTransformer} builds upon the inverted embedding structure and tries to capture dependencies between multiple variables. TimeXer~\cite{TimeXer} integrates endogenous and exogenous information by leveraging patch-wise self-attention and variate-wise cross-attention simultaneously. Even though the performance is promising, the scalability of these methods is limited on high-dimensional datasets due to the complexity of attention mechanism.

\section{Methodology}
\label{sec:method}

\subsection{Problem definition and Overview} 
In MTSF, given historical observations \(X = \{x_1, \ldots, x_T\} \in \mathbb{R}^{T \times N}\), where \(T\) represents the number of time steps and \(N\) represents the number of variables, our target is predicting the future \(S\) time steps \(Y = \{x_{T+1}, \ldots, x_{T+S}\} \in \mathbb{R}^{S \times N}\).

\begin{figure*}[htbp]
  \centering
  \includegraphics[width=0.85\textwidth]{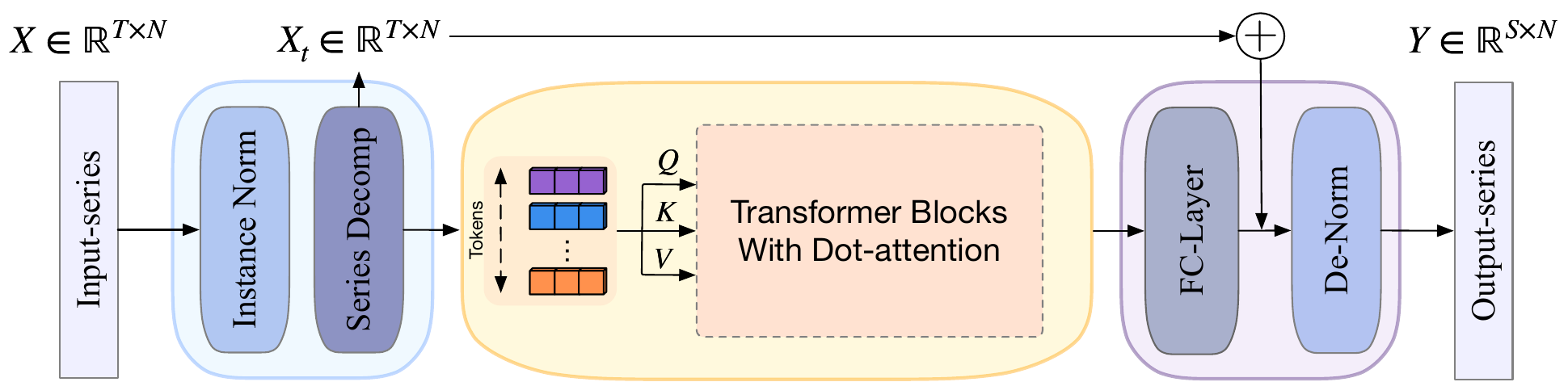}
  \caption{Overall structure of \textbf{Ister}. The original sequence is partitioned into seasonal and trend components using the $SeriesDecomp(.)$. After obtaining the embedded representations of the seasonal and trend components, the seasonal component is fed into the Transformer, resulting in the seasonal outputs. The seasonal outputs and trend part are added together after the projection to obtain the final result.}
  \label{fig:model}
\end{figure*}

\noindent \textbf{Overview.} In this section, we present the overall architecture of the proposed Ister (Figure \ref{fig:model}). First, we describe the proposed Dot-attention Mechanism in Section 3.2. Then, we introduce backbone architecture and computational details of Ister in Section 3.3. The overall pipeline of Ister is shown in Algorithm~\ref{alg:Ister}.

\begin{algorithm}[h]
\caption{Ister - Overall Architecture.}
\label{alg:Ister}
\textbf{Input}: Lookback time series $X\in\mathbb{R}^{T\times N}$.\\
\textbf{Parameter}: input Length $T$; predicted length $S$; variates number $N$; token dimension $D$; Transformer block number $L$.\\
\textbf{Output}: Future time series $\hat{Y}\in\mathbb{R}^{S\times N}$.

\begin{algorithmic}[1] 
\STATE $X_s, X_t = SeriesDecomp(X)$
\STATE $X_s, X_t = X_s.\texttt{transpose}, X_t.\texttt{transpose}$
\STATE $H^{0} = \texttt{Embedding}(X_s)$
\FOR{$l \in \{1,\dots,L\}$}
    \STATE $H^{l}=\texttt{Encoder}_{s}\big(H^{l-1}\big)$
\ENDFOR
\STATE $\hat{Y_s} = \texttt{MLP}(H^{L})$
\STATE $\hat{Y_t} = \texttt{MLP}(X_t)$
\STATE $\hat{Y} = \hat{Y_s} + \hat{Y_t}$
\STATE \textbf{Return} $\hat{Y}$
\end{algorithmic}
\end{algorithm}

\subsection{Dot-attention Mechanism}
Previous studies have highlighted the importance of capturing inter-series dependencies to improve prediction accuracy~\cite{iTransformer}. Existing methods rely on the self-attention mechanism to model these dependencies. However, in practical applications, data typically involve a large number of channels, and the quadratic computational complexity of self-attention makes it impractical for long-sequence processing. At the same time, many channels in time-series contain significant noise or lack semantic meaning, making it challenging to manually distinguish the semantic information of each channel.

In order to solve the aforementioned bottleneck, while improving both efficiency and accuracy, we propose \textit{Dot-attention} mechanism. Dot-attention eliminates the multi-head architecture and replaces the matrix multiplication in self-attention with element-wise multiplication. The formula for Dot-attention is as follows:

\begin{equation}
\begin{aligned}
& \text{\textit{Dot.}}(Q, K, V) = \left(\sum_{i=1}^{N} \text{Softmax}(Q_{i}) \odot K_{i}\right)^T \mathbf{1}_N^T \odot V, \\
 & \text{where } Q, K, V \in \mathbb{R}^{N \times D} 
\end{aligned}
\end{equation}

The Dot-attention mechanism, similar to the original multi-head attention mechanism, consists of three learnable weight matrices: 1. \( W_{Q} \): models the importance of different channels and assigns weights to channel information. 2. \( W_{K} \): aggregates the weighted representations of all features into a global predictive representation. 3. \( W_{V} \): using the global predictive representation to guide the predictions for all channels. Specifically, the input tokens are first multiplied by \( W_{Q} \), \( W_{K} \), and \( W_{V} \) to obtain the corresponding representations: \( Q \), \( K \), and \( V \). A \textit{softmax} operation is then applied to the weight representations, converting each channel's values into a probability distribution used to scale each channel's contribution. The weight representation is then multiplied by the global representation via a dot product to extract the core representation of the sequence. Finally, this core representation is multiplied by \( V \), which serves as the guide for each channel prediction.

All computational operations of Dot-attention consist of element-wise multiplications. This not only reduces the original quadratic computational complexity of self-attention to \(O(N)\) but also enhances accuracy. Table \ref{tab:attention_complexities} lists the time complexities of the existing Transformers. The Dot-attention used by Ister achieves optimal computational efficiency and has the lowest complexity.

\begin{table}[htbp]
\centering
\caption{Computational complexities of various algorithms.}
\begin{tabular}{>{\raggedright\arraybackslash}p{1.5cm} >{\raggedright\arraybackslash}p{3.5cm} > {\raggedright\arraybackslash}p{2cm} }
\toprule
\textbf{Model} & \textbf{Attention Algorithm} & \textbf{Complexity} \\
\midrule
Ister & Dot-attention & $O(N)$\\
Transformer & Multi-head attention & $O(N^2)$\\
Reformer & LSH attention & $O(N \cdot \log N)$\\
Informer & ProbSparse & $O(N \cdot \log N)$\\
Autoformer & Auto-Correlation & $O(N \cdot \log N)$\\
FEDformer & FEA-f & $O(N)$\\
\bottomrule
\end{tabular}
\label{tab:attention_complexities}
\end{table}

\subsection{Method Details}

\noindent \textbf{Instance Normalization.} Distribution shifts between training and testing datasets are common in time series data~\cite{Non-station}. RevIN~\cite{Revin} has shown that applying simple instance normalization strategies between the model input and output can effectively mitigate this issue. We use RevIN to eliminate the non-stationary statistics in the input sequence.

\noindent \textbf{Series Decomposition.} 
Time series can be decomposed into seasonal and trend components using decomposition algorithm. Seasonal components preserve the periodic features, while trend components depict the overall fluctuations. We use a moving average technique to smooth periodic fluctuations and emphasize long-term trends. For an input series $X \in \mathbb{R}^{T \times N}$ with $T$ time steps, the decomposition process is as follows:
\begin{equation}
X_t = F(X),
X_s = X - X_t
\end{equation}
where $F(.)$ is an average pooling filter, $X_s$ and $X_t$ $\in \mathbb{R}^{T \times N}$ represent the seasonal and extracted trend-cyclical components of $X$, respectively. This process is encapsulated by $X_s, X_t = SeriesDecomp(X)$, which serves as an integral component within the model architecture.

After seasonal-trend decomposition, Ister extract the seasonal components from the input. Then we utilize attention to process seasonal component.

\noindent \textbf{Transformer Backbone.} Transformer-based method has great ability in capturing long-range dependencies, thus it works well for seasonal component, which encompasses the most significant features of time-series. Thus, we use Transformer to predict seasonal component.

\begin{equation}
\begin{aligned}
    H_s^0 &= \text{Embedding}_s(X_s), \\
    H_s^l &= \text{Encoder}_s(H_s^{l-1}), \quad l = 1, \ldots, L, \\
    \hat{Y}_s &= \text{Projection}_s(H_s^L), \\
\end{aligned}
\end{equation}

where $H \in \mathbb{R}^{N \times D}$ contains $N$ embedded tokens of dimension $D$ and the superscript denotes the layer. We implement Embedding$_s$ and Projection$_s$ using MLP network. The obtained seasonal variate tokens are independently processed by a shared feed-forward network.

\section{Experiment}

\noindent \textbf{Datasets.} We extensively include 11 real-world datasets in our experiments, including Electricity, ETT (4 subsets), Traffic, Weather~\cite{autoformer} and PEMS (4 subsets)~\cite{SCINet}.


\noindent \textbf{Implementation details.} Our method is trained with L2 loss, using the ADAM optimizer \cite{adam}. For the rest of the parameter settings, we strictly follow the settings of TimeXer \cite{TimeXer}. For each of the models, we explore the number of Transformer blocks $L$ within the set \{1, 2, 3, 4\}, and the dimension of series $D$ within \{128, 256, 512\}. All experiments are repeated three times, implemented in PyTorch \cite{pytorch} and conducted on NVIDIA RTX 3090 24GB GPUs.

\subsection{Forecasting Results}

\noindent \textbf{Baselines.}  We carefully choose competitive forecasting models in MTSF task as our benchmark, including S-Mamba~\cite{S-Mamba}, TimeXer~\cite{TimeXer}, iTransformer~\cite{iTransformer}, MSGNet~\cite{MSGNet}, TimesNet~\cite{TimesNet}, PatchTST~\cite{PatchTST}, Crossformer~\cite{Crossformer}, DLinear~\cite{DLinear} and SCINet~\cite{SCINet}.

\noindent \textbf{Results.}  Comprehensive forecasting results are listed in Table \ref{tab:main_mean} with best in \textbf{bold} and the second best is \underline{underlined}. The lower MSE/MAE indicates the more accurate prediction result. Across 22 prediction tasks, Ister achieved top-2 performance in 20 of them. Ister is particularly good at forecasting high-dimensional time series, such as the Traffic dataset (862 variables) and the PEMS dataset (170 to 883 variables). Moreover, our model demonstrates the most significant improvements over iTransformer~\cite{iTransformer} at most forecasting lengths, with some datasets even showing enhanced predictive performance as the forecast horizon increases.

\begin{table*}[!htb]
\centering
\caption{Comparison of multivariate time series forecasting results across 11 real-world datasets. We set the prediction lengths $S$ to \{12, 24, 48, 96\} for PEMS dataset and \{96, 192, 336, 720\} for others. The look-back length \(T\) is uniformly fixed at 96. The best results are highlighted in \textbf{bold}, the second best are \underline{underlined}, and the \textit{Count} row counts the number of times each model ranks in the top 2.}
\label{tab:main_mean}
\begin{adjustbox}{max width=\linewidth}
\begin{tabular}{@{}c|cc|cc|cc|cc|cc|cc|cc|cc|cc|cc@{}}
\toprule
Model & \multicolumn{2}{c|}{\begin{tabular}[c]{@{}c@{}}Ister\\ (Ours) \end{tabular}} & \multicolumn{2}{c|}{\begin{tabular}[c]{@{}c@{}}S-Mamba\\ (2025) \end{tabular}} & \multicolumn{2}{c|}{\begin{tabular}[c]{@{}c@{}}TimeXer\\ (2024) \end{tabular}} & \multicolumn{2}{c|}{\begin{tabular}[c]{@{}c@{}}iTransformer\\ (2024) \end{tabular}} & \multicolumn{2}{c|}{\begin{tabular}[c]{@{}c@{}}MSGNet\\ (2024) \end{tabular}} & \multicolumn{2}{c|}{\begin{tabular}[c]{@{}c@{}}TimesNet\\ (2023) \end{tabular}} & \multicolumn{2}{c|}{\begin{tabular}[c]{@{}c@{}}PatchTST\\ (2023) \end{tabular}} & \multicolumn{2}{c|}{\begin{tabular}[c]{@{}c@{}}Crossformer\\ (2023) \end{tabular}} & \multicolumn{2}{c|}{\begin{tabular}[c]{@{}c@{}}DLinear\\ (2023) \end{tabular}} & \multicolumn{2}{c}{\begin{tabular}[c]{@{}c@{}}SCINet\\ (2022) \end{tabular}} \\ \midrule
Metric & MSE & MAE & MSE & MAE & MSE & MAE & MSE & MAE & MSE & MAE & MSE & MAE & MSE & MAE & MSE & MAE & MSE & MAE & MSE & MAE \\ \midrule
ETTh1 & \underline{0.438} & \underline{0.438} & 0.455 & 0.450 & \textbf{0.437} & \textbf{0.437} & 0.454 & 0.448 & 0.453 & 0.453 & 0.458 & 0.450 & 0.469 & 0.455 & 0.529 & 0.522 & 0.456 & 0.452 & 0.747 & 0.647 \\
ETTh2 & \textbf{0.349} & \textbf{0.387} & 0.381 & 0.405 & 0.368 & 0.396 & 0.383 & 0.407 & 0.413 & 0.427 & 0.414 & 0.427 & 0.387 & 0.407 & 0.942 & 0.684 & 0.559 & 0.515 & 0.954 & 0.723 \\
ETTm1 & \underline{0.386} & \underline{0.399} & 0.398 & 0.405 & \textbf{0.382} & \textbf{0.397} & 0.407 & 0.410 & 0.400 & 0.412 & 0.400 & 0.406 & 0.387 & 0.400 & 0.513 & 0.495 & 0.403 & 0.407 & 0.486 & 0.481 \\
ETTm2 & \underline{0.279} & \underline{0.325} & 0.288 & 0.332 & \textbf{0.274} & \textbf{0.322} & 0.288 & 0.332 & 0.289 & 0.330 & 0.291 & 0.333 & 0.281 & 0.326 & 0.757 & 0.611 & 0.350 & 0.401 & 0.571 & 0.537 \\
Electricity & \textbf{0.167} & \textbf{0.260} & \underline{0.170} & \underline{0.265} & 0.171 & 0.270 & 0.178 & 0.270 & 0.194 & 0.301 & 0.193 & 0.295 & 0.205 & 0.290 & 0.244 & 0.334 & 0.212 & 0.300 & 0.571 & 0.537 \\
Traffic & \textbf{0.399} & \textbf{0.270} & 0.414 & \underline{0.276} & 0.466 & 0.287 & 0.428 & 0.282 & 0.660 & 0.382 & 0.620 & 0.336 & 0.481 & 0.300 & 0.550 & 0.304 & 0.625 & 0.383 & 0.804 & 0.509 \\
Weather & \underline{0.243} & \textbf{0.271} & 0.251 & 0.276 & \textbf{0.241} & \textbf{0.271} & 0.258 & 0.278 & 0.249 & 0.278 & 0.259 & 0.287 & 0.259 & 0.273 & 0.259 & 0.315 & 0.265 & 0.317 & 0.292 & 0.363 \\
PEMS03 & \textbf{0.108} & \underline{0.217} & 0.122 & 0.228 & \underline{0.112} & \textbf{0.214} & 0.113 & 0.222 & 0.150 & 0.251 & 0.147 & 0.248 & 0.180 & 0.291 & 0.169 & 0.282 & 0.278 & 0.375 & 0.114 & 0.224 \\
PEMS04 & 0.106 & 0.213 & \underline{0.103} & \underline{0.211} & 0.105 & 0.209 & 0.111 & 0.221 & 0.122 & 0.239 & 0.129 & 0.241 & 0.195 & 0.307 & 0.209 & 0.314 & 0.295 & 0.388 & \textbf{0.093} & \textbf{0.202} \\
PEMS07 & \underline{0.092} & \underline{0.193} & \textbf{0.089} & \textbf{0.188} & 0.085 & 0.182 & 0.101 & 0.204 & 0.122 & 0.227 & 0.125 & 0.226 & 0.211 & 0.303 & 0.235 & 0.315 & 0.329 & 0.396 & 0.119 & 0.217 \\
PEMS08 & \textbf{0.136} & \underline{0.226} & 0.148 & \textbf{0.224} & 0.175 & 0.250 & \underline{0.150} & \underline{0.226} & 0.205 & 0.285 & 0.193 & 0.271 & 0.280 & 0.321 & 0.268 & 0.307 & 0.379 & 0.416 & 0.159 & 0.244 \\ \midrule
Count & \multicolumn{2}{c|}{\textbf{20}} & \multicolumn{2}{c|}{8} & \multicolumn{2}{c|}{\underline{10}} & \multicolumn{2}{c|}{2} & \multicolumn{2}{c|}{0} & \multicolumn{2}{c|}{0} & \multicolumn{2}{c|}{0} & \multicolumn{2}{c|}{0} & \multicolumn{2}{c|}{0} & \multicolumn{2}{c}{2} \\ \bottomrule
\end{tabular}
\end{adjustbox}
\end{table*}

\subsection{Model Analysis}

\begin{table}[h]
    \centering
    \caption{Ablations on Ister. Results are averaged from all prediction length. For ETT dataset, the results are averaged from four datasets \{ETTm1, ETTm2, ETTh1, ETTh2\}.}
    \resizebox{0.48\textwidth}{!}{
    \begin{tabular}{c|c|c|c|c}
        \toprule[1.0pt]
        Models & ETT & ECL & Traffic & Weather\\
        \cmidrule(lr){2-5}
        Metric & MSE ~~ MAE & MSE ~~ MAE & MSE ~~ MAE & MSE ~~ MAE\\
        \midrule[0.5pt]
        w/o $Dot.$ & 0.432 ~~ 0.447 & 0.199 ~~ 0.281 & 0.499 ~~ 0.319 & 0.265 ~~ 0.283\\
        \midrule[0.5pt]
        $+ MSA$ & 0.368 ~~ 0.390 & 0.174 ~~ 0.268 & 0.430 ~~ 0.285 & 0.248 ~~ 0.276\\
        \midrule[0.5pt]
        w/o periodicity & 0.411 ~~ 0.405 & 0.179 ~~ 0.287 & 0.405 ~~ 0.273 & 0.248 ~~ 0.275\\
        \midrule[0.5pt]
        Ister (Ours) & \textbf{0.363} ~~ \textbf{0.387} & \textbf{0.168} ~~ \textbf{0.260} & \textbf{0.399} ~~ \textbf{0.270} & \textbf{0.243} ~~ \textbf{0.271}\\
        \bottomrule[1.0pt]
        \end{tabular}
        }
    \label{tab:ab}
\end{table}

\noindent \textbf{Ablation study.} We conducted ablation experiments on four datasets to evaluate the contribution of different components in Ister to the overall performance improvement. Specifically, 'w/o $Dot.$' represents the removal of all attention modules in the Ister, '$+ MSA$' means that we replace the Dot-attention modules with Multi-head self attention, 'w/o periodicity' indicate the removal of series decomposition module. The ablation results are shown in Table \ref{tab:ab}.


The key findings are twofold. First, extracting seasonal components significantly enhances predictive performance, demonstrating that channel-wise alignment benefits substantially from explicit modeling of periodic patterns. Second, replacing standard multi-head self-attention with our proposed dot-attention mechanism yields consistent performance gains across datasets, underscoring the effectiveness and efficiency of the designed attention scheme.

\begin{figure}[htbp]
\centering
\includegraphics[width=0.48\textwidth]{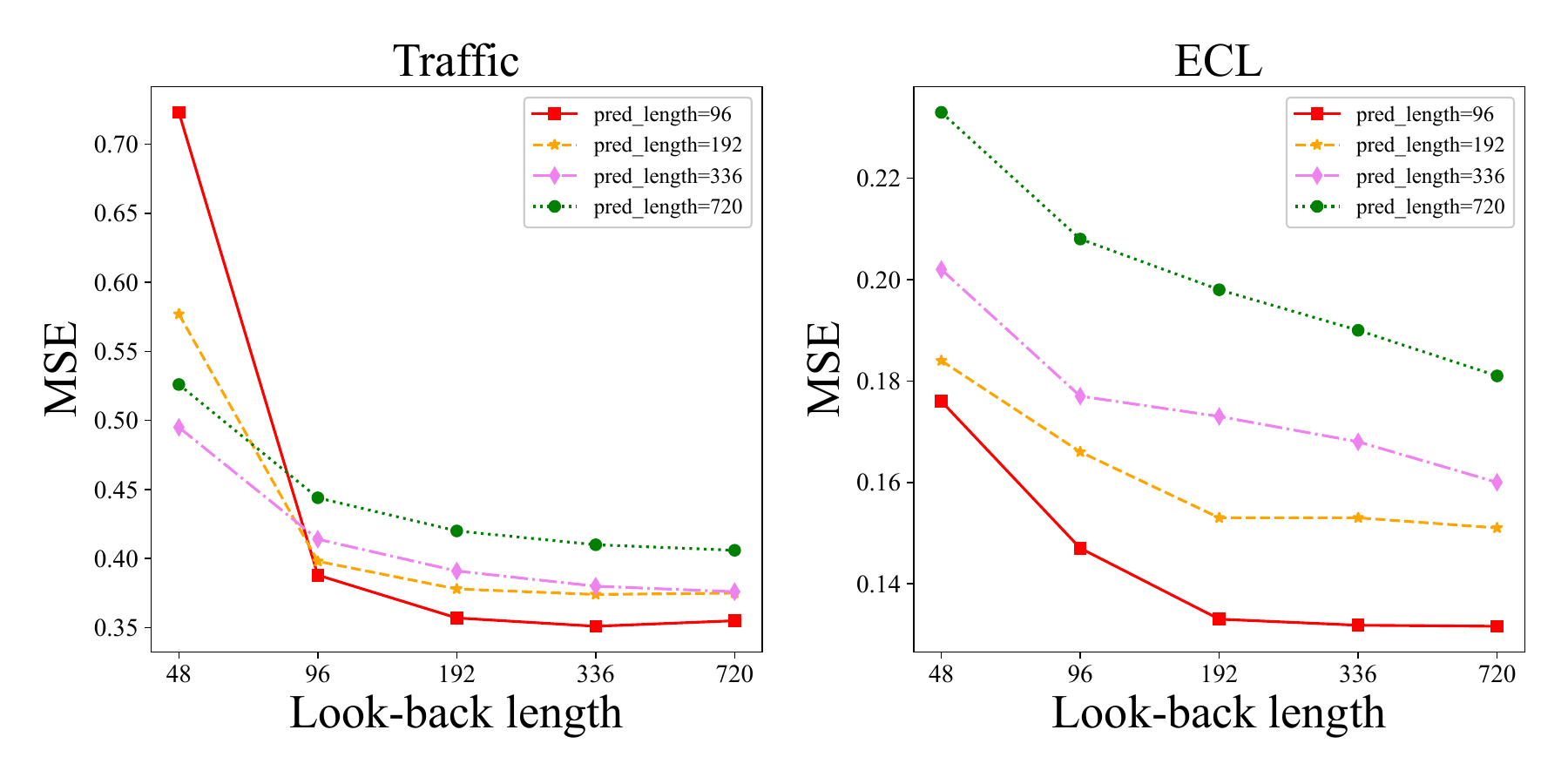}
\caption{Forecasting performance with the look-back length varying from \{48, 96, 192, 336, 720\} and prediction length varying from \{96, 192, 336, 720\}. Ister's forecasting performance benefits from the increase of look-back length.}
\label{lookback}
\end{figure}

\noindent \textbf{Increasing lookback length.} Previous works have shown that the forecasting performance does not improve with the increase in lookback length on Transformers~\cite{DLinear}. However, as historical information expands, our model can better learn the temporal dependencies, achieving performance improvement across multiple prediction steps with increasing lookback window size. As shown in Figure \ref{lookback}, the MSE monotonically decreases with increasing length of the lookback window at any prediction length. 

\begin{table}[h]
\centering
\caption{Comparison of practical efficiency of LTSF-Transformers under input length=96 and predict length=720 on the Traffic. The training time averages 5 runs.}
\begin{tabular}{c|ccc} \hline
Method & Parameter & Time & Memory\\\hline
Ister  & \textbf{7.670M} & \textbf{0.098s} &\textbf{3924MiB} \\\hline
Informer & 14.39M & 0.154s & 4107MiB\\
FEDformer & 21.21M & 0.332s & 4720MiB\\
PatchTST & 9.170M & 0.125s & 6761MiB\\
Crossformer & 72.37M & 0.393s & 11430MiB \\\hline        
\end{tabular}
\label{tab:efficiency}
\end{table}

\noindent \textbf{Model efficiency.} Since Dot-attention mechanism has linear time complexity, the Ister's overall complexity is \( O(N) \). For memory complexity, we embed the whole series of dimension \(T\) as the tokens of dimension \(D\). This makes our model's memory usage less dependent on the prediction horizon compared to other Transformer-based methods, where memory usage increases linearly with the prediction horizon. Consequently, as shown in the Table \ref{tab:efficiency}, the proposed Ister achieves better efficiency for long-term sequence predictions.

\noindent \textbf{Generalization performance.} 
We evaluated the generalization performance of \textit{Dot-attention} on other models by selecting three different types of models utilizing self-attention: iTransformer (Channel-wise), PatchTST (Patch-wise), and Transformer (Point-wise). Experiments were conducted on three high-dimensional multivariate datasets: ECL (321 channels), Traffic (862 channels), and PEMS08 (170 channels). Table \ref{table:dot result} presents the experimental results and the corresponding percentage improvements. Our findings indicate that \textit{Dot-attention}, which is designed for inter-channel attention, performs well in other tasks as well. It not only retains most of the performance but also achieves improvements on certain datasets. The Dot-attention module can be seamlessly integrated into any channel independence Transformer-based model, enhancing predictive performance.

\begin{table}[h]
\centering
\caption{We compared the performance of original \textit{Multi-head self-attention} and \textit{Dot-attention} for several different types (Channel-wise, Patch-wise and Point-wise) of Transformers.}
\resizebox{0.48\textwidth}{!}{
\begin{tabular}{c|c|cc|cc|cc}
\toprule
\multicolumn{2}{c|}{Models} & \multicolumn{2}{c|}{iTransformer} & \multicolumn{2}{c|}{PatchTST} & \multicolumn{2}{c}{Transformer} \\
\cmidrule(lr){3-4} \cmidrule(lr){5-6} \cmidrule(lr){7-8}
\multicolumn{2}{c|}{Metric} & MSE & MAE & MSE & MAE & MSE & MAE \\
\midrule[0.5pt]
\multirow{3}{*}{\rotatebox{90}{ECL}} & $Ori.$ & 0.179 & 0.269 & 0.204 & 0.290 & 0.414 & 0.476\\
& +$Dot.$ & \textbf{0.172} & \textbf{0.264} & \textbf{0.194} & \textbf{0.288} & \textbf{0.269} & \textbf{0.360}\\
\cmidrule(lr){2-8}
& \textit{Promotion} & 4.0\% & 2.0\% & 5.0\% & 1.0\% & 35\% & 25\% \\
\midrule[0.5pt]
\multirow{3}{*}{\rotatebox{90}{Traffic}} & $Ori.$ & 0.412 & 0.281 & 0.502 & 0.323 & 0.660 & 0.361\\
& +$Dot.$ & \textbf{0.404} & \textbf{0.272} & \textbf{0.450} & \textbf{0.291} & \textbf{0.645} & \textbf{0.359}\\
\cmidrule(lr){2-8}
& \textit{Promotion} & 4.1\% & 4.3\% & 11\% & 10\% & 2.3\% & 0.6\% \\
\midrule[0.5pt]
\multirow{3}{*}{\rotatebox{90}{PEMS08}} & $Ori.$ & 0.175 & 0.254 & 0.280 & 0.321 & 0.252 & 0.274\\
&+$Dot.$ & \textbf{0.169} & \textbf{0.250} & \textbf{0.194} & \textbf{0.263} & \textbf{0.241} & \textbf{0.263}\\
\cmidrule(lr){2-8}
& \textit{Promotion} & 3.5\% & 1.6\% & 31\% & 19\% & 4.5\% & 4.0\% \\
\bottomrule
\end{tabular}
}
\label{table:dot result}
\end{table}

\section{Conclusion}
In this paper, we investigate the task of multivariate time series forecasting (MTSF), which is a pressing need for real-world applications. To address the scalability issues of existing Transformer-based forecasters, we propose \textit{Ister}, an efficient linear Transformer model which improving the predictive performance by capturing the inter-series dependencies. The proposed \textit{Dot-attention} mechanism improves computational efficiency and predictive accuracy. We conducted experiments on real-world datasets and showed that Ister achieves state-of-the-art performance on almost all datasets, highlighting Ister as an efficient solution for MTSF tasks.

\clearpage
\newpage

\bibliographystyle{IEEEbib}
\bibliography{strings,refs}

\end{document}